\documentclass[nohyperref]{article}

\usepackage{microtype}
\usepackage{graphicx}
\usepackage{booktabs} 

\usepackage{hyperref}

\usepackage[accepted]{icml2022}

\usepackage{amsmath}
\usepackage{amssymb}
\usepackage{mathtools}
\usepackage{amsthm}

\usepackage[capitalize,noabbrev]{cleveref}

\theoremstyle{plain}

\theoremstyle{definition}

\theoremstyle{remark}

\usepackage[textsize=tiny]{todonotes}

\usepackage[utf8]{inputenc} 

\usepackage{amsmath,amsfonts,bm}

\def\eqref#1{equation~\ref{#1}}

\def\1{\bm{1}}

\def\ro{{\textnormal{o}}}

\def\rs{{\textnormal{s}}}

\def\vb{{\bm{b}}}

\def\vk{{\bm{k}}}

\def\vq{{\bm{q}}}

\def\vv{{\bm{v}}}

\def\vx{{\bm{x}}}
\def\vy{{\bm{y}}}

\def\evx{{x}}
\def\evy{{y}}

\def\mI{{\bm{I}}}

\def\mK{{\bm{K}}}

\def\mQ{{\bm{Q}}}

\def\mS{{\bm{S}}}

\def\mV{{\bm{V}}}
\def\mW{{\bm{W}}}

\DeclareMathAlphabet{\mathsfit}{\encodingdefault}{\sfdefault}{m}{sl}
\SetMathAlphabet{\mathsfit}{bold}{\encodingdefault}{\sfdefault}{bx}{n}

\def\gW{{\mathcal{W}}}

\newcommand{\E}{\mathbb{E}}

\newcommand{\R}{\mathbb{R}}

\newcommand{\softmax}{\mathrm{softmax}}

\usepackage[T1]{fontenc}    
\usepackage{url}            
\usepackage{amsfonts}       
\usepackage{nicefrac}       
\usepackage{xcolor}         
\usepackage{times}
\usepackage{latexsym}
\usepackage{tikz}
\usepackage{multirow}
\usepackage{subcaption}
\usepackage{adjustbox}
\usepackage{xspace}
\usetikzlibrary{fit,positioning,decorations.pathmorphing,calc}

\usepackage{makecell}
\usepackage{pifont}
\newcommand{\cmark}{\ding{51}}%
\newcommand{\xmark}{\ding{55}}%

\usepackage{wrapfig}
\usepackage{changepage}
\usepackage{eucal,amsbsy}
\usepackage[normalem]{ulem}
\usepackage{tablefootnote}

\newcommand\numberthis{\addtocounter{equation}{1}\tag{\theequation}}

\newenvironment{itemizesquish}{\begin{list}{\labelitemi}{\setlength{\itemsep}{-0.2em}\setlength{\labelwidth}{0.5em}\setlength{\leftmargin}{\labelwidth}\addtolength{\leftmargin}{\labelsep}}}{\end{list}}

\newcommand{\imagenet}{\texttt{ImageNet1k}\xspace}
\newcommand{\cifar}{\texttt{CIFAR-100}\xspace}
\newcommand{\coco}{\texttt{COCO}\xspace}
\newcommand{\model}{\textsc{ripple}\xspace}

\definecolor{strings}{rgb}{.674,.251,.259}
\definecolor{keywords}{rgb}{.224,.451,.686}
\definecolor{comment}{rgb}{.422,.451,.322}

\newcommand{\appendixhead}{\textbf{\huge Appendices}\vspace{0.2in}}

\icmltitlerunning{Ripple Attention for Visual Perception with Sub-quadratic Complexity}

\begin{document}

\twocolumn[
\icmltitle{Ripple Attention for Visual Perception with Sub-quadratic Complexity}

\begin{icmlauthorlist}
\icmlauthor{Lin Zheng}{dept}
\icmlauthor{Huijie Pan}{dept}
\icmlauthor{Lingpeng Kong}{dept,lab}
\end{icmlauthorlist}

\icmlaffiliation{dept}{Department of Computer Science, The University of Hong Kong}
\icmlaffiliation{lab}{Shanghai Artificial Intelligence Laboratory}

\icmlcorrespondingauthor{Lin Zheng}{linzheng@connect.hku.hk}

\icmlkeywords{Machine Learning, ICML}

\vskip 0.3in
]

\printAffiliationsAndNotice{}  

\begin{abstract}

Transformer architectures are now central to sequence modeling tasks. At its heart is the attention mechanism, which enables effective modeling of long-term dependencies in a sequence. 
Recently, transformers have been successfully applied in the computer vision domain, where 2D images are first segmented into patches and then treated as 1D sequences. Such linearization, however, impairs the notion of spatial locality in images, which bears important visual clues. To bridge the gap, we propose \emph{ripple attention}, a sub-quadratic attention mechanism for vision transformers. Built upon the recent kernel-based efficient attention mechanisms, we design a novel dynamic programming algorithm that weights contributions of different tokens to a query with respect to their relative spatial distances in the 2D space in linear observed time.
Extensive experiments and analyses demonstrate the effectiveness of ripple attention on various visual tasks.
\end{abstract}

\section{Introduction}
\label{sec:intro}

The transformer architecture \citep{vaswani2017attention} has been dominant in various important natural language processing (NLP) tasks, including machine translation \citep{vaswani2017attention,dehghani2018universal}, language understanding \citep{devlin2018bert}, language modeling \citep{dai-etal-2019-transformer,baevski2018adaptive} and many others. The cornerstone of a transformer is the attention mechanism \citep{bahdanau2014neural} which computes pair-wise interactions between any token pairs of the input sequence. As a result, it is capable of modeling long-term dependencies in a sequence, which is an important factor to 
the success of transformers.

Recently, the transformer architecture has also found its applications in the domain of computer vision (CV). It is adopted for image classification \citep{vit,deit,swin-transformer, focal, pvtv2}, segmentation \citep{instance-segmentation,segmenter}, low-level image processing \citep{IPT}, image generation \citep{igpt}, object detection \citep{detr,cond-detr} and many other tasks. In these vision applications, a 2D image is represented as a set of patches flattened into a 1D sequence. These patches are analogous to the tokens in sequence modeling tasks that are commonly seen in NLP. Nevertheless, such linearization undermines the inherent local structure of a 2D image, which bears important visual clues \citep{cnn-inductive-bias}. There often exist strong correlations within local neighborhoods in an image. Therefore, paying more attention to patches in a closer region could facilitate gathering information that is particularly useful in visual pattern recognition. This is similar to the concept of \emph{context} in NLP, just that the structural context of a visual token is scattered in the 1D sequence, making it difficult for the transformer to capture such prior knowledge. In contrast, the convolutional neural network (CNN) \citep{cnn1,cnn2,cnn3}, which has been the de-facto architecture in computer vision tasks for decades, utilizes local receptive fields and achieves good performance. The drawback of that is, as convolution operations are limited to small receptive fields, they have great difficulty in extracting global image features.  

Therefore, it is appealing to incorporate the notion of spatial vicinity into the transformer, while still preserving its capacity of modeling long-term dependencies. To bridge the gap, we propose 
\emph{ripple attention} (Figure~\ref{fig:ripple-diagram}; \S\ref{sec:model}), an efficient attention mechanism for vision transformers based on recently proposed linearized attention variants (\S\ref{ssec:rfa}). In ripple attention, contributions from different tokens to a query are weighted with respect to their relative spatial distances in the 2D space. These spatial weights are derived through a stick-breaking transformation (\S\ref{ssec:spatial_weights}), which promotes local correlations by leaning to assign larger weights to spatially closer tokens. We then design a dynamic programming algorithm (\S\ref{ssec:dp}) that is capable of executing ripple attention in linear observed time, taking advantage of the recently proposed linearized attention (\S\ref{ssec:rfa}) and the summed-area table technique (\S\ref{ssec:dp}).
 
We validate our method by conducting extensive experiments on image classification and object detection tasks (\S\ref{sec:ex}). Ripple attention significantly improves the accuracy of the original vision transformer in image classification and performs competitively with detection transformers for object detection (\S\ref{ssec:main-ex}), in asymptotically faster runtime (\S\ref{ssec:time}). Further analysis on the rippling distance and ablation studies (\S\ref{ssec:ripple-distance}) indicate that ripple attention favors contributions from tokens in the vicinity yet preserves global information from long-term dependencies.

\section{Preliminary}
\label{sec:prelim}

\subsection{Attention Mechanism}
\label{ssec:attn}
Let $\mQ \in \R^{N \times D}$ denote a set of $N$ query vectors, which attend to $M$ key and value vectors, denoted by matrices $\mK \in \R^{M \times D}$ and $\mV \in \R^{M \times C}$ respectively. For a query vector at position $n$, the softmax attention function computes the following quantity\footnote{We omit the scaling factor for simplicity.}:
\begin{align*}
  \operatorname{Attn}\left(\vq_{n},\mK,\mV\right)&=\sum_{m=1}^M \frac{\exp \left(\vq_{n}^\top \vk_{m} \right)}{\sum_{m'=1}^M \exp \left(\vq_{n}^\top \vk_{m'} \right)} \vv_{m}^{\top}\\
  &\coloneqq \softmax(\mK\vq_{n} )^{\top}\mV \numberthis{}\label{eqn:attn},
\end{align*}
which is an average of the set of value vectors $\mV$ weighted by normalized similarity between different queries and keys. However, such quantity requires computing the similarity between all pairs of queries and keys, incurring quadratic complexity in both time and memory. It makes the computational overhead for long sequences prohibitive, especially in the case of vision tasks.

\subsection{Linearized Attention}
\label{ssec:rfa}
To reduce the computational complexity in attention mechanism, prior works propose to linearize the softmax kernel \citep{performer,katharopoulos2020transformers,rfa}. In particular, they replace the exponential kernel used in softmax functions $\kappa(\vq,\vk) \coloneqq \exp \left(\vq^\top \vk \right)$ with a dot product of two feature maps $\phi(\vq)^\top \phi(\vk)$, where $\phi(\cdot) : \R^D \rightarrow \R^{D'}$. Further details about the choice of feature maps can be found in Appendix \ref{app:ssec:rfa}. With the feature map, linearized attention can be written as:
\begin{align*}
  \operatorname{LA}\left(\vq_{n},\mK,\mV\right) &\coloneqq
  \sum_{m=1}^M \frac{\phi(\vq_{n})^\top\phi(\vk_{m})}{\sum_{m'=1}^M \phi(\vq_{n})^\top\phi(\vk_{m'})} \vv_{m}^{\top}\\
  &= \frac{\phi(\vq_{n})^\top\sum_{m=1}^M\phi(\vk_{m})\vv_{m}^{\top}}{\phi(\vq_{n})^\top\sum_{m'=1}^M \phi(\vk_{m'})} \numberthis{}\label{eqn:rfa}.
\end{align*}
In other words, by grouping together the computations of keys and values, their statistics can be shared for all queries. It therefore achieves linear complexity in both time and memory with respect to the length of the sequence, as we only need to compute $\sum_{m=1}^M\phi(\vk_{m})\vv_{m}^{\top}$ and $\sum_{m=1}^M\phi(\vk_{m})$ once and then reuse them for each query.

\section{Model}
\label{sec:model}

In this section, we introduce ripple attention, a novel attention mechanism that features the relative spatial vicinity. We start from a reformulation of the linearized attention (\S\ref{ssec:rfa}) under the notation of vicinal groups (\S\ref{ssec:ripple-attn}). This reformulation makes it straightforward to introduce a spatial weight associated with each vicinal group, which is the cornerstone of ripple attention. We then describe the derivation of these spatial weights through a stick-breaking transformation (\S\ref{ssec:spatial_weights}) and a dynamic programming algorithm (\S\ref{ssec:dp}) based on the summed-area table technique to perform the computation of ripple attention efficiently.

\subsection{Ripple Attention}
\label{ssec:ripple-attn}
\begin{figure*}
\centering
\begin{subfigure}[b]{0.280\textwidth} 
  \centering
  {\includegraphics[width=0.8\textwidth]{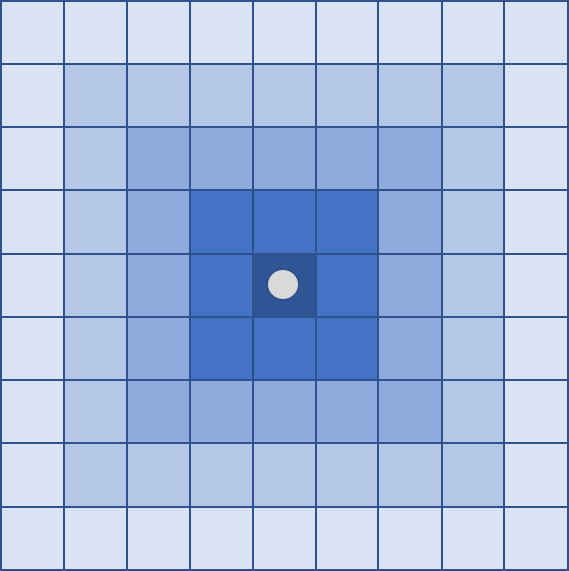}}
  \caption{}
  \label{fig:ripple-a}
\end{subfigure}
\begin{subfigure}[b]{0.280\textwidth} 
  \centering
  {\includegraphics[width=0.8\textwidth]{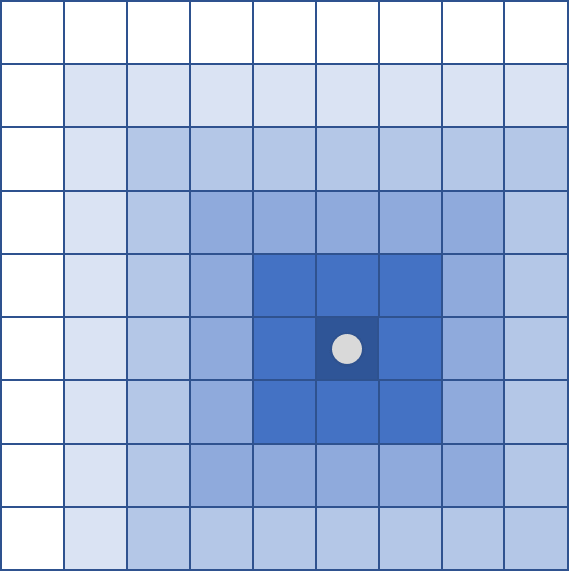}}
    \caption{}
    \label{fig:ripple-b}
\end{subfigure}
\begin{subfigure}[b]{0.280\textwidth} 
  \centering
  {\includegraphics[width=0.8\textwidth]{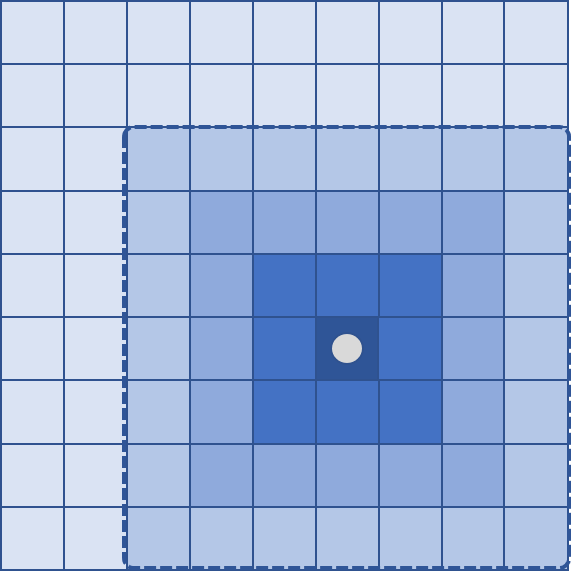}}
    \caption{}
    \label{fig:ripple-c}
\end{subfigure}
\caption{A demonstration of vicinal groups in ripple attention on a $9 \times 9$ image $\mI$. Each square denotes a token, the circle denotes the query position and we use deeper color to indicate a larger spatial weight. \textbf{Left (a)}: an example of vicinal group partitioning in the case where the query lies in the center of an image, resulting in a symmetric rippling effect over the 2D space; \textbf{Middle (b)}: another group partitioning on the same image but the query token is not centered. In this case, distal vicinal groups (the top-left corner) receives almost no spatial weights. \textbf{Right (c)}: the same case as in (b) with the threshold $\tau$ set such that $\hat{r}=4$. Groups beyond $\hat{r} - 1$ (indicated by the dashed line) contribute according to an equal spatial weight.}
\label{fig:ripple-diagram}
\end{figure*}

We assume an input image consists of $H \times W$ patch tokens. Given a query token at position $(i,j)$, we partition the whole set of patch tokens into $R + 1$ vicinal groups $\{\mathcal{N}_r(i,j)\}_{r=0}^R$, according to their Chebyshev (or chessboard) distances $r$ from the position to the query, which means for every token at position $(m,n) \in \mathcal{N}_r(i,j)$ we have $\max(|m - i|, |n-j|) = r$. 
Illustrations of such vicinal group partitioning can be found in Figure~\ref{fig:ripple-a} or Figure~\ref{fig:ripple-b}, where each group is marked by a different color.

Under the notation of vicinal groups, we can reformulate the linearized attention $\operatorname{LA}\left(\vq_{ij},\mK,\mV\right)$ as:
\begin{align*}
  \frac
  {\phi(\vq_{ij})^\top\sum_{r=0}^R \sum_{(m,n) \in \mathcal{N}_r(i,j)}\phi(\vk_{mn})\vv_{mn}^{\top}}
  {\phi(\vq_{ij})^\top\sum_{r=0}^R \sum_{(m',n') \in \mathcal{N}_r(i,j)} \phi(\vk_{m'n'})} \numberthis{}\label{eqn:rfa_groups}.
\end{align*}
This formulation is computationally equivalent since the summations over $\phi(\vk)\vv^{\top}$ and $\phi(\vk)$ also cover all positions within the image. The essence of ripple attention is to let tokens respond differently to a query, according to their relative distances. Typically, tokens close to the query in the 2D space should weigh more than the tokens far away in general, since there exist strong local correlations in images. This control is translated into a spatial weight $\alpha_r(i,j)$ associated with each vicinal group $\mathcal{N}_r(i,j)$ and can be easily introduced into linearized attention\footnote{The partitioning is hard to implement in softmax attention mechanism. More discussions about ripple-softmax and its complexity can be found in Appendix~\ref{app:sec:complexity}.}:
\begin{align*}
  &\operatorname{RippleAttn}\left(\vq_{ij},\mK,\mV\right) \coloneqq \\
  &\frac
  {\phi(\vq_{ij})^\top\sum_{r=0}^R \alpha_{r}(i,j)\sum_{(m,n) \in \mathcal{N}_r(i,j)}\phi(\vk_{mn})\vv_{mn}^{\top}}
  {\phi(\vq_{ij})^\top\sum_{r=0}^R \alpha_{r}(i,j)\sum_{(m',n') \in \mathcal{N}_r(i,j)} \phi(\vk_{m'n'})} \numberthis{}\label{eqn:ripple}.
\end{align*}
We define $\alpha_r(i,j) \in (0,1) ,\forall r=0,1,\dots,R$ and $\sum_{r = 0}^R \alpha_r(i,j) = 1$, to reweigh contributions of different vicinal groups with respect to a query in the attention computation. By respecting the spatial structure of images, ripple attention constructs a structured context for queries, which facilitates the model to reconcile both global and local information.
The name \textit{ripple attention} comes from its similarity to the ripples on the surface of water (Figure~\ref{fig:ripple-diagram}).

\subsection{Spatial Weights}
\label{ssec:spatial_weights}
To derive the spatial weights $\alpha_r \in (0,1) \,\,\forall r=0,1,\dots,R$,\footnote{In this section, we sometimes drop the dependence on position $(i,j)$ for spatial weights $\alpha_r(i,j)$ when there is no ambiguity.} we first define a sequence of scalars $\{\rs_r\}$, where $\rs_r \in  (0,1) \,\forall r=1,\dots,R$. The spatial weights are parameterized as follows (with $\rs_{R+1} = 1$):
\begin{equation}
  \alpha_r =
  \begin{cases}
  \rs_{1}, &\text { if } r=0 \\
  \rs_{r+1} \prod_{r' \leq r}\left(1-\rs_{r'}\right), &\text { otherwise }
  \end{cases}
  \label{eqn:sbt}
\end{equation}
The sequence $\{\rs_r\}$ is generated through a small neural network followed by a sigmoid function. See Appendix~\ref{app:ssec:spatial_weights_param} for more details. Our construction is analogous to the stick-breaking process in Bayesian nonparametrics \citep{wasserman2006all}, except that $\rs_r$ here is a deterministic scalar rather than a random variable. One of its appealing properties is:
\begin{align*}
    &\sup \left\{\rs_{r+1}\prod_{i \leq r}\left(1-\rs_{i}\right): \rs_{r+1} \in (0,1)\right\} \geq \\
    &\quad\sup \left\{\rs_{r'+1}\prod_{i \leq r'}\left(1-\rs_{i}\right): \rs_{r'+1} \in (0,1)\right\} \text{ if } r < r'.
\end{align*}
In other words, we only assume the \emph{supremum} of each spatial weight $\alpha_r$ is decreasing. A stronger constraint could be monotonicity, for example $\alpha_r > \alpha_{r'}$ if $r < r'$. We argue that the former is more favorable, because it offers the flexibility to let distant tokens outweigh when necessary; the effective modeling of such long-term dependencies is deemed as the key to the success of the transformer architecture as well.

In theory, the stick-breaking transformation produces $R + 1$ different spatial weights. However, as the weights become trivially small near the end of this transformation, the computational overhead incurred by them becomes worthless. Therefore, we define a threshold $\tau$ to adaptively terminate the transformation at vicinal group $\mathcal{N}_{\hat{r}}(i,j)$ when the length of remaining stick is less than $\tau$ (i.e., $1 - \sum_{r = 0}^{\hat{r}} \alpha_{r}(i,j) < \tau$). We then merge all vicinal groups $\mathcal{N}_{r}(i,j)$ with $r \geq \hat{r}$ and share the same weights among them assuming they contribute equally:
\begin{equation*}
\alpha_{r}(i,j) = \frac{1 - \sum_{r' = 0}^{\hat{r}-1} \alpha_{r'}(i,j)}{R-\hat{r}+2}, \text{ if } r >= \hat{r}.\numberthis{}\label{eqn:adaptive_weights}
\end{equation*}
This truncating-and-merging operation is demonstrated in Figure~\ref{fig:ripple-c}, where the threshold $\tau$ is set such that $\hat{r} = 4$. In this case, all the remaining groups (outside the dashed line) share the same weight according to \eqref{eqn:adaptive_weights}. Compared to Figure~\ref{fig:ripple-b}, adaptive ripple allows a stop of the transformation before hitting the boundary, which prevents potentially worthless computations for distal groups. At the same time, it does weigh in the contributions from those groups, preserving the ability to capture long-term dependencies.

\subsection{Dynamic Programming}
\label{ssec:dp}
The only problem left now is how to compute ripple attention effectively. A na\"ive implementation of \eqref{eqn:ripple} has a time complexity of $\mathcal{O}(HWR^2)$. Since $R$ is bounded by $\max(H, W) - 1$, the computation is quadratic with respect to the length of the sequence. We give detailed derivations of runtime complexity of each attention variant in Appendix~\ref{app:sec:complexity}.

In this section, we present a dynamic programming algorithm built on the summed-area table (SAT) technique, a classic algorithm in computer graphics and computer vision \citep{crow1984summed,viola2001robust}, which reduces the time complexity of ripple attention to $\mathcal{O}(HWR)$. SAT is an efficient data structure that stores prefix sums for each pixel position of an image such that summations over any window in the image can be retrieved in constant time. For an image $\mI$ with height $H$ and width $W$, it first initializes the table by computing the cumulative sum $\mS$ of all the tokens above and to the left of $(i,j)$ inclusively in the 2D plane\footnote{In practice, we adopt a linear-complexity implementation which first performs the cumulative summation over the row axis and then the column axis, yielding the same result.}:
\begin{equation}
  \mS(i,j) = \sum_{i' = 1}^i \sum_{j' = 1}^j \mI(i', j') \label{eqn:sat}.
\end{equation}
For a square window with center $(i,j)$ and radius $r$ (i.e., a region centered at $(i,j)$ with both its height and width equal to $2r + 1$), the summation over its elements is denoted by $\mathcal{W}(i,j,r)$ and can be computed in constant time \citep{crow1984summed}:
\begin{align*}
  &\mathcal{W}(i,j,r) \coloneqq \mS(i+r, j+r) - \mS(i-r-1,j+r) - \\
  &\quad\mS(i+r, j-r-1) + \mS(i-r-1, j-r-1) \numberthis{}\label{eqn:window-sum}.
\end{align*}
In this work, we consider $\{\phi(\vk_{ij})\vv^{\top}_{ij}\}$ and $\{\phi(\vk_{ij})\}$ as generalized pixels within the input image and construct two SATs $\mS_1$ and $\mS_2$ to compute their prefix summations respectively. According to \eqref{eqn:window-sum}, the window sums can be obtained efficiently and are denoted as $\mathcal{W}_1$ and $\mathcal{W}_2$.

We show that the sum of $\phi(\vk)\vv^{\top}$ and $\phi(\vk)$ over vicinal group $\mathcal{N}_r(i,j)$ can also be computed within constant time from SATs:
\begin{align}
  \sum_{(m,n) \in \mathcal{N}_r(i,j)}\!\!\phi(\vk_{mn})\vv_{mn}^{\top} &= \mathcal{W}_1(i,j,r) - \mathcal{W}_1(i,j,r-1);\nonumber\\
  \sum_{(m,n) \in \mathcal{N}_r(i,j)}\phi(\vk_{mn}) &= \mathcal{W}_2(i,j,r) - \mathcal{W}_2(i,j,r-1).\nonumber
\end{align}
Intuitively, this can be viewed as taking the difference between the largest square window wrapped by the group and the smallest square window containing the group.
Equipped with SATs, the formulation of ripple attention becomes:
\begin{align*}
  &\operatorname{RippleAttn}\left(\vq_{ij},\mK,\mV\right) 
  = \\
  &\frac{\phi(\vq_{ij})^\top\sum_{r=0}^R \alpha_{r}(i,j)\left(\mathcal{W}_1(i,j,r) - \mathcal{W}_1(i,j,r-1)\right) }{\phi(\vq_{ij})^\top\sum_{r=0}^R \alpha_{r}(i,j) \left(\mathcal{W}_2(i,j,r) - \mathcal{W}_2(i,j,r-1)\right)}. \numberthis{}\label{eqn:ripple_sat}
\end{align*}
In \S\ref{ssec:spatial_weights}, we merge all groups $\mathcal{N}_{r}(i,j)$ with $r >= \hat{r}$ assuming equal contributions (equation~\ref{eqn:adaptive_weights}), which can be jointly computed in constant time using $\mS(H,W) - \mathcal{W}(i,j,\hat{r} - 1)$. Therefore, given a reasonable hyper-parameter choice of $\tau$, the algorithm can achieve linear observed time in the sequence length. This is due to the fact that after the precomputation of SATs (in linear complexity), for each query the required summations of vicinal groups can be computed in constant time.

\paragraph{Efficient Gradient Computation.}
The algorithm discussed above addresses the runtime complexity of the forward pass of ripple attention. In Appendix~\ref{app:sec:gradient}, we also present a dynamic programming algorithm to compute gradients for the backward pass, again in $\mathcal{O}(HWR)$ time and space complexity. The main idea is to utilize the symmetry of vicinal groups and reformulate the gradient calculations as summations over different groups, where computations could be further reduced using SATs; in contrast, a nai\"ve implementation would come with $\mathcal{O}(HWR^2)$ complexity.

\paragraph{Complexity Analysis.}
As mentioned above, ripple attention runs in $\mathcal{O}(HWR)$ time complexity with the help of dynamic programming on the introduced vicinal groups; and it could achieve linear observed runtime in practice with appropriate hyper-parameter configuration. Algorithm \ref{alg:ripple} sketches the dynamic programming for the ripple attention mechanism given a single query and a threshold $\tau$.\footnote{Note that the algorithm can be easily executed in parallel for all queries.} Due to the flexibility of ripple attention, its time complexity can be further improved if we adapt the step size of the rippling process. For example, we could achieve $\mathcal{O}(HW\log R)$ time complexity if we allow ripples to be exponentially thicker; see \S\ref{ssec:log_rippling} for more detailed discussion.
As for the memory consumption, we observe that the tensor $\mathcal{W}$ does not even need to be explicitly materialized, since previously computed results for closer vicinal groups could be reused for more distant vicinal groups. Therefore, the space complexity of ripple attention remains $\mathcal{O}(HW)$ irrespective of the rippling distance $R$.

\begin{algorithm}[t]
\caption{Dynamic Programming for Ripple Attention}
\label{alg:ripple}
\begin{algorithmic}
\STATE \textbf{Input:} the key-value statistics $I \in \R^{H \times W}$, query position $(i,j)$, spatial weights $\{\alpha_r(i,j)\}$ and threshold $\tau$;
\STATE \textbf{Output:} The weighted sum over vicinal groups $\texttt{res} \coloneqq \sum_{r=0}^R \alpha_{r}(i,j)\left(\mathcal{W}(i,j,r) - \mathcal{W}(i,j,r-1)\right)$;
\STATE
\STATE \textbf{Initialize} $\texttt{res} \gets 0$, \, $\gW_\text{cur} \gets 0$, \, $\gW_\text{prev} \gets 0$, \, $S \in \R^{H \times W} \gets \mathbf{0}$;
\STATE \textbf{Compute} summed-area table $S$ by calling $\texttt{cumsum()}$ function twice over horizontal and vertical directions respectively;
\STATE \textbf{Compute} $\hat{r}$ by summing over $\alpha_{r}(i,j)$ until $1-\sum_{r=0}^{\hat{r}}\alpha_{r}(i,j) < \tau$;
\FOR{$r = 0, 1, \dots, \hat{r} - 1$}
\STATE {\fontsize{8.4}{8.4}\selectfont \(\triangleright\) \texttt{Cumulative sum at the bottom-right corner}}
\STATE $\mS_1 \gets \mS(i+r,j+r)$;
\STATE {\fontsize{8.4}{8.4}\selectfont \(\triangleright\) \texttt{Cumulative sum at the bottom-left corner}}
\STATE $\mS_2 \gets \mS(i-r-1,j+r)$;
\STATE {\fontsize{8.4}{8.4}\selectfont \(\triangleright\) \texttt{Cumulative sum at the top-right corner}}
\STATE $\mS_3 \gets \mS(i+r, j-r-1)$; 
\STATE {\fontsize{8.4}{8.4}\selectfont \(\triangleright\) \texttt{Cumulative sum at the top-left corner}}
\STATE $\mS_4 \gets \mS(i-r-1, j-r-1)$;
\STATE {}
\STATE {\fontsize{8.4}{8.4}\selectfont \(\triangleright\) \texttt{Compute sum over elements in the window}}
\STATE $\gW_\text{cur} \gets \mS_1 - \mS_2 - \mS_3 + \mS_4$;
\STATE $\texttt{res} \gets \texttt{res} + \alpha_r(i,j) (\gW_\text{cur} - \gW_\text{prev})$;
\STATE $\gW_\text{prev} \gets \gW_\text{cur}$;
\ENDFOR
\STATE {\fontsize{8.4}{8.4}\selectfont \(\triangleright\) \texttt{Compute the sum over all remaining groups}}
\STATE $\texttt{res} \gets \texttt{res} + \alpha_{\hat{r}}(i,j) (\mS(H,W) - \mathcal{W}(i,j,\hat{r} - 1))$;
\STATE \textbf{return} \texttt{res}
\end{algorithmic}
\end{algorithm}

\section{Experiments}
\label{sec:ex}
We conduct extensive experiments on image classification and detection tasks to demonstrate the effectiveness of ripple attention.

\subsection{Experimental Setup}
\label{ssec:setup}
\paragraph{Datasets} For image classification, we evaluate our model on standard benchmark datasets: (1) \imagenet dataset \citep{imagenet}, consisting of approximately 1,280K/50K images of 1000 classes for training/validation splits respectively; (2) \cifar \citep{cifar}, which contains 50K images of 100 classes for training and 10K for evaluation. For detection tasks, we conduct our experiment on the \coco benchmark \citep{coco} consisting of 118k training and 5k validation images respectively. 

\paragraph{Baselines}

Our model for image classification is based on the vision transformer architecture (ViT) \citep{vit,deit}, where the attention block is replaced with ripple attention. We compare ripple attention (referred to as \model hereafter) with various attention mechanisms in ViT: 
\begin{itemizesquish}
    \item \textsc{deit}, which adopts the same architecture as ViT and vanilla softmax attention.\footnote{To facilitate comparisons and simplify experimental settings, we do not use the distillation technique.}
    \item \textsc{convit}, which imposes a soft convolutional inductive bias on the vanilla attention mechanism.
    \item \textsc{deit-la}, a \textsc{deit} model equipped with linearized attention (\S\ref{ssec:rfa}) instead of softmax attention. We also include several variants that improve \textsc{deit-la}, such as \textsc{permuteformer} \citep{permuteformer}, \textsc{spe} \citep{spe} and Rotary positional embeddings \citep[\textsc{rope},][]{rotary} that incorporates relative positional encodings.
\end{itemizesquish}

For object detection, we evaluate our model in the general framework of detection transformer \citep[DETR;][]{detr} to test the generalization ability of \model. However, due to the slow convergence of \textsc{detr}, we are unable to run \textsc{detr} model for a full training schedule given limited computational resources. Instead, we adopt \textsc{smca} \citep{smca} as our baseline, a variant of \textsc{detr} that greatly speeds up the convergence by constraining the attention map in the decoder side. Our model, referred to as \textsc{smca-ripple}, replaces all attention blocks with ripple attention in the transformer encoder. For completeness, we also compare with \textsc{smca-la}, an \textsc{smca} variant that adopts linearized attention in encoder attention block.

\subsection{Main Implementation Details}
Here we discuss key ingredients for implementing \model; see Appendix~\ref{app:sec:implementation} for more comprehensive implementation details and discussions.

\paragraph{Feature Map Parameterization.}
Note that \model is based on the linearized attention mechanism (\S\ref{ssec:rfa}). In this work, the feature map $\phi(\cdot)$ is defined to be deterministic with learnable parameters, which consists of a two-layer MLP with trigonometric and ReLU activations in turn. We find it works well in our experiments. Detailed discussions about our choice and a corresponding ablation study can be found in Appendix~\ref{app:ssec:rfa}.

\paragraph{Rippling Attention Specifications.}
In \S\ref{ssec:spatial_weights} we define a threshold $\tau$ to control the termination of rippling process. In practice we find it beneficial to introduce a hard constraint $R_{\text{max}}$ such that the model explicitly limits the maximum distance of rippling propagation to $R_{\text{max}}$ and then merges all the remaining groups. In this way, we could not only further reduce the computation overhead, but also encourage the attention mechanism to allocate more weights to distal groups. This can be seen as a stronger version of halting threshold $\tau$, which is easier to tune due to a more intuitive effect on the rippling process. Given $R_{\text{max}}$, our model is robust to the change of $\tau$; therefore, we set $\tau$ to 0.001 throughout our experiments and only conduct ablation studies on $R_{\text{max}}$ (\S\ref{ssec:ripple-distance}). We find an intermediate value gives a reasonable trade-off between local and long-term dependencies. 

Furthermore, when applied to vision transformers for image classification tasks, we only replace the first several attention layers with ripple attention, while the remaining ones adopt linearized attention.

\subsection{Main Results}
\label{ssec:main-ex}
\paragraph{Results on ImageNet-1K Dataset.} The results of comparisons among \model and other models on \imagenet dataset are presented in Table~\ref{tb:imagenet-acc}. 
We observe \model outperforms both \textsc{deit-la}, upon which \model is built, and its variants by a large margin. Although \textsc{deit-la} gives a clear performance drop compared to the standard vision transformer \textsc{deit}, \model still performs better than \textsc{deit} and achieves results comparable to the improved variant \textsc{convit} while in asymptotically faster runtime, which clearly demonstrates the effectiveness of our approach.
\begin{table}
	\caption{Image classification results for different vision transformers on \imagenet dataset. All the variants of \textsc{deit-la}, including \textsc{permuteformer}, are trained by us. }
	\label{tb:imagenet-acc}
	\centering
	\resizebox{\columnwidth}{!}{ 
		\begin{tabular}{c | c | c | c}
			\toprule[.1em]
			 {Model} & {\# Params} & {Top-1 Acc.} & {Top-5 Acc.} \\
            \midrule
    \multicolumn{4}{c}{Models with quadratic complexity} \\
      \midrule
			\textsc{deit} & 5.72M & 72.20 & 91.10\\
			\textsc{convit} \citep{convit} & 5.72M & \textbf{73.11} & \textbf{91.71}\\
            \midrule
      \multicolumn{4}{c}{Models with sub-quadratic complexity} \\
      \midrule
			\textsc{deit-la} & 5.76M & 70.67 & 90.16 \\
			\textsc{deit-la} + \textsc{sincspe} \citep{spe} & 5.84M & 67.32 & 88.14 \\
			\textsc{deit-la} + \textsc{convspe} \citep{spe} & 6.69M & 67.64 & 88.40\\
			\textsc{deit-la} + \textsc{rope} \citep{rotary} & 5.76M & 71.19 & 90.48 \\
			\textsc{permuteformer} \citep{permuteformer} & 5.76M & 71.42 & 90.51\\
            \midrule
            \model & 5.78M  & 73.02 & 91.56\\
		\bottomrule[.1em]
	\end{tabular}}
	\vskip -0.15in
\end{table}

\paragraph{Results on CIFAR-100 Dataset.} 
We further conduct experiments on \cifar dataset and report results in Table~\ref{tb:cifar100-acc}. \model outperforms both \textsc{deit-la} and \textsc{deit} by a substantial margin on \cifar dataset, and also achieves competitive performance compared to \textsc{convit}. This suggests that \model also generalizes well on a relatively smaller dataset.
Following the setting in \citep{cvt, contnet}, we also make a comparison of these models in the absence of absolute positional embeddings. We observe a significantly larger performance gap between vanilla vision transformers and models designed to incorporate the notion of locality (Table~\ref{tb:cifar100-acc}). This implies \model could structure the scattered spatial context, which is beneficial to information aggregation among patches. Still, the performance decrease in \model in the absence of positional embeddings suggests that absolute global positions contain complementary information to the prior knowledge of locality, which is also consistent with a recent study \citep{ape1}.
\begin{table}[t]
	\caption{Classification results for different vision transformers on \cifar dataset. All of these models are trained by us. APE denotes absolute positional embeddings.}
	\label{tb:cifar100-acc}
	\centering
	\resizebox{\columnwidth}{!}{    
		\begin{tabular}{c | c  c  c || c  c  c}
			\toprule[.1em]
			\multirow{2.5}{*}{Model} & \multicolumn{3}{c||}{w/ APE} & \multicolumn{3}{c}{w/o APE} \\
			\cmidrule{2-7}
			 & {\# Params} & {Top-1 Acc.} & {Top-5 Acc.} & {\# Params} & {Top-1 Acc.} & {Top-5 Acc.}\\
      \midrule
            \textsc{deit-la}  & 5.42M & 67.00 & 88.57 & 5.36M & 54.04 & 79.66\\
			\textsc{deit}  & 5.42M & 67.87 & 89.71 & 5.36M & 53.64 & 80.30\\
			\textsc{convit}  & 5.42M & \textbf{74.34} & \textbf{92.87} & 5.36M & \textbf{73.88} & \textbf{92.20}\\
		\midrule
		    \model & 5.47M & 73.94 & 92.37 & 5.42M  & 72.94 & 91.86\\
		\bottomrule[.1em]
	\end{tabular}
	}  
\end{table}

\paragraph{Results on COCO Benchmark.} In Table~\ref{tb:coco-ap} we report the results for object detection. Again, we see the same trend that the performance drops by over 2 AP when using linearized attention in the encoder (\textsc{smca-la}). However, \textsc{smca-ripple} improves \textsc{smca-la} on all object scales with a marginal increase of GFLOPs and almost catches up with \textsc{smca}. The mAP gap between \textsc{smca-ripple} and \textsc{smca} is further narrowed down from 0.5 to 0.3 with 108 training epochs. In addition, \textsc{smca-ripple} achieves better results than \textsc{smca} on small scale objects, which is attributed to the promoted locality of ripple attention.
\begin{table}
	\caption{Object detection results for different detection transformers on \coco benchmark under both 50 training epoch schedule and 108 epoch schedule.}
	\label{tb:coco-ap}
	\centering
	\resizebox{\columnwidth}{!}{    
		\begin{tabular}{c  c c  c | c c  c  c || c  c  c  c}
			\toprule[.1em]
			\multirow{2.5}{*}{Model} &  \multirow{2.5}{*}{\# Params} & \multirow{2.5}{*}{GFLOPs} &  \multirow{2.5}{*}{\makecell[c]{Inference\\time(s)}} & \multicolumn{4}{c||}{50 epochs} & \multicolumn{4}{c}{108 epochs} \\
			\cmidrule{5-12}
			& & & & {AP} & {AP\tiny{S}} & {AP\tiny{M}} & {AP\tiny{L}} &  {AP} & {AP\tiny{S}} & {AP\tiny{M}} & {AP\tiny{L}} \\
            \midrule
			\textsc{smca} & 41.5M & 88& 0.059 & \textbf{41.0}  & 21.9 & \textbf{44.3} & \textbf{59.1} & \textbf{42.7} & 22.8 & \textbf{46.1} & \textbf{60.0} \\
			\textsc{smca-la} & 41.7M & 79 & 0.062 & 39.1  & 19.8 & 42.8 & 56.5 & 41.1 & 22.0 & 44.5 & 59.0\\
			\textsc{smca-ripple}  & 41.8M & 80& 0.065 & 40.5 & \textbf{22.1} & 44.1 & 57.7 & 42.3 & \textbf{23.2} & 45.6 & \textbf{60.0}\\
      \midrule[.1em]
	\end{tabular}
	}  
	\vskip -0.15in
\end{table}

\section{Analysis}
\label{sec:analysis}

\begin{figure*}[t]
\centering
\begin{subfigure}{0.40\textwidth} 
  \centering
  {\includegraphics[width=\textwidth]{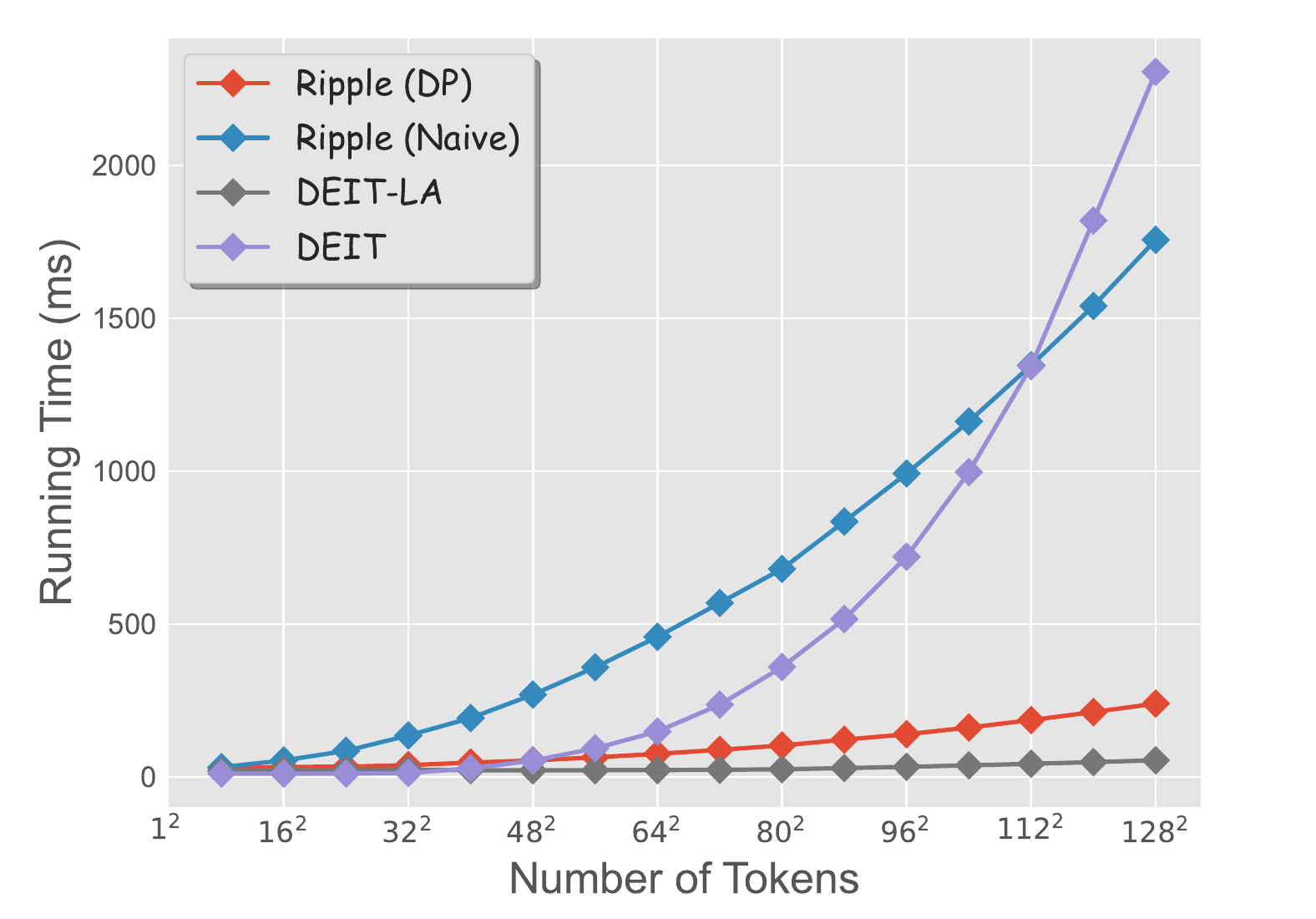}}
\end{subfigure}
\begin{subfigure}{0.40\textwidth} 
  \centering
  {\includegraphics[width=\textwidth]{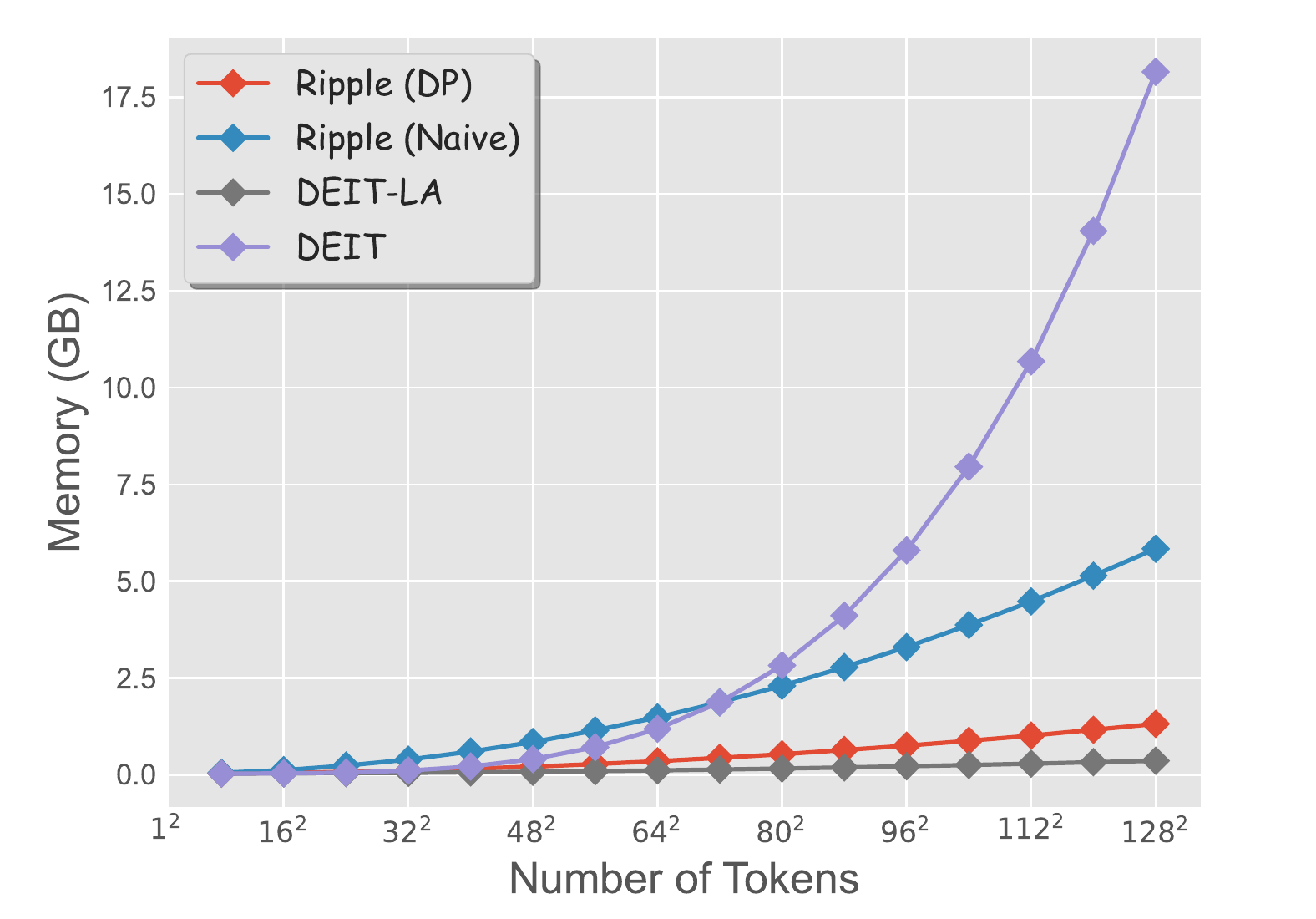}}
\end{subfigure}
\caption{Empirical running time (left) and memory consumption (right) under different numbers of tokens. All models are tested with a batch size of 4 on a single NVIDIA V100 GPU machine, averaged by 10 runs.}
\label{fig:ripple-speed}
\vskip -0.1in
\end{figure*}

\subsection{Inspecting the Rippling Process}
\label{ssec:ripple-distance}

\begin{table}[t]
	\caption{Classification results on \cifar dataset under different setups of the rippling process. The speed is measured by the number of images processed per second with a batch size of 64. \model w/o \textsc{sbt} represents ripple attention whose spatial weights are generated through a softmax function instead of stick breaking transforms (\textsc{sbt}). ``--'' indicates not applicable.}
	\label{tb:analysis-distance}
	\centering
	\resizebox{\columnwidth}{!}{    
		\begin{tabular}{c | c | c | c | c | c | c}
\toprule[.1em]
Model &  locality & global dep.  & $R_{\text{max}}$ & Speed & Top-1 Acc. & Top-5 Acc. \\
\midrule
\multirow{4}{*}{\model} & \multirow{4}{*}{\cmark} & \multirow{4}{*}{\cmark}
& 2  & 832 & 72.65  & 91.83  \\
& & & 4  & 792 & \textbf{73.94}  & \textbf{92.37}  \\
& & & 8  & 578 & 73.37  & 92.21  \\
& & & 16 & 382 & 73.48  & 92.25  \\
\midrule
\textsc{fixed-ripple} & \cmark & \cmark & 4 & 795 & 71.34 & 90.77 \\
\midrule
\textsc{logarithmic-ripple} & \cmark & \cmark & -- & 810 & 73.05 & 91.78 \\ 
\midrule
\textsc{truncated-ripple} & \cmark & \xmark & 4 & 820 & 72.18 & 91.66\\
\midrule
\model w/o \textsc{sbt} & \xmark & \cmark & 4 & 784 & 71.94 & 91.83\\
\midrule
\textsc{deit-la} & \xmark & \cmark & --  & 2664 & 67.00 & 88.57  \\
\bottomrule[.1em]
	\end{tabular}}  
	\vskip -0.15in
\end{table}
\paragraph{On the Effect of Maximum Rippling Distances.}
The maximum rippling distance $R_{\text{max}}$ defined in \S\ref{ssec:setup} controls the boundary of \model with informative spatial weights. 
To evaluate its effect on the modeling performance, we vary the maximum rippling distance $R_{\text{max}}$ and report the results on \cifar dataset in Table~\ref{tb:analysis-distance}. Overall, \model performs well with a moderate or larger $R_{\text{max}}$. If $R_{\text{max}}$ is too small, the performance drops significantly, although still outperforming \textsc{deit-la}. It can be attributed to the fact that if the stick-breaking transformation terminates early, the query would attend mostly to its immediate spatial neighbors while not sufficiently respecting global dependencies. 

\paragraph{On the Effect of Stick Breaking Transforms.} We construct a variant of \model with fixed and exponentially-decayed spatial weights (i.e., $1/2, (1/2)^2, (1/2)^3, \cdots, (1/2)^{R_{\text{max}}}$), which is denoted by \textsc{fixed-ripple}. This variant appears to be a trivial solution for incorporating the locality into the model, although it does not respect potentially strong long-term dependencies and only assign diminishing weights to distal groups. We find \model with hard-coded weights also performs better than \textsc{deit-la}, which indicates the effectiveness of recovering spatial structures in transformers. Our stick-breaking transformation gives a further boost over the fixed-weight baseline, thanks to its flexibility in the spatial weights. We also consider a baseline \model w/o \textsc{sbt}, which replaces the stick-breaking transformation (\S\ref{ssec:spatial_weights}) with a simple softmax function to generate spatial weights. This variant does not promote any locality but has the potential to learn such pattern implicitly. It performs slightly better than hard-coded weights and much worse compared to \model, verifying the effectiveness of stick-breaking transformation.

\begin{figure}[t]
\centering
\includegraphics[width=0.6\columnwidth]{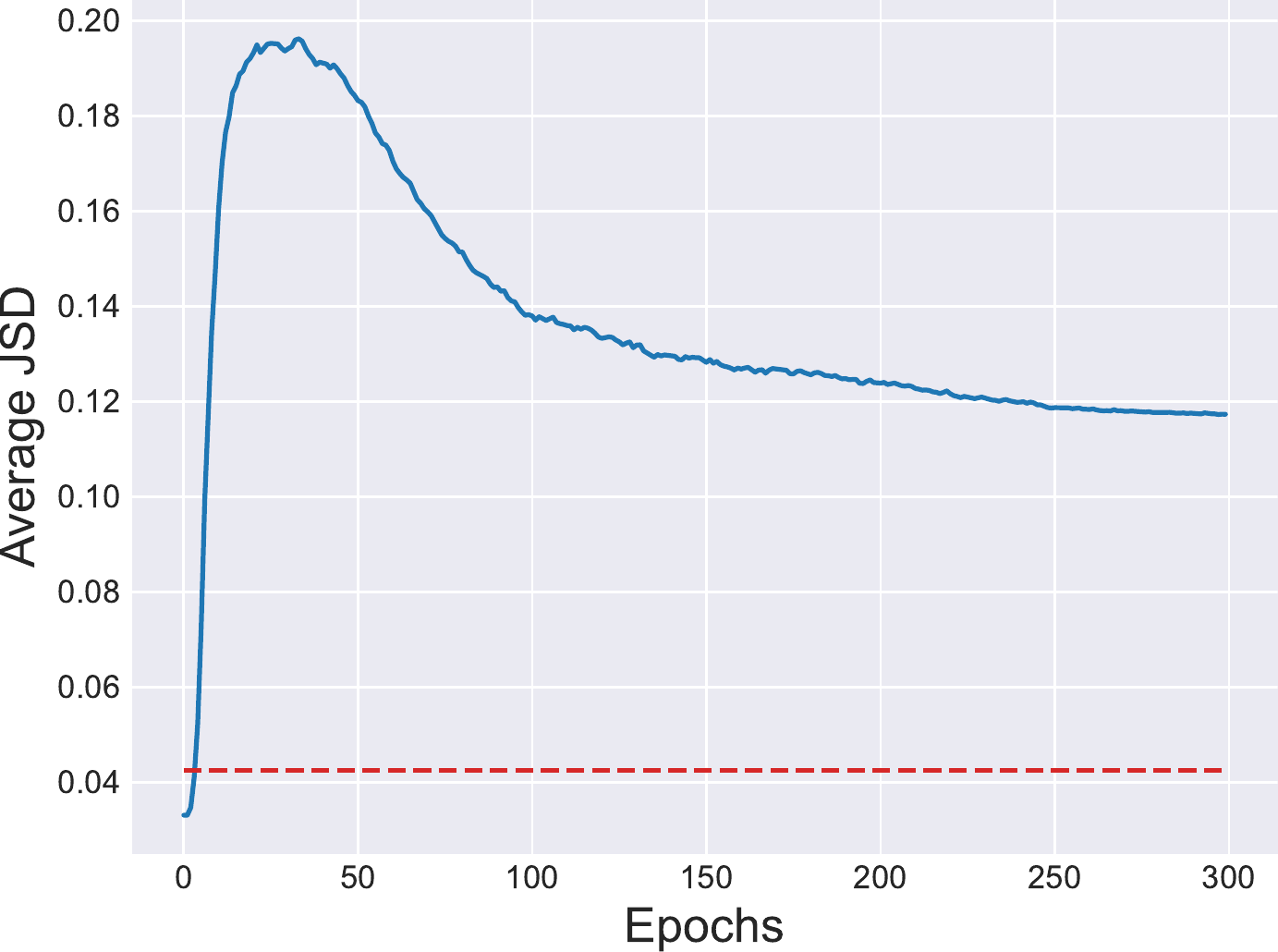}
\caption{Training dynamic of average JSD between induced distributions of the proposed stick-breaking transform (SBT) and fixed exponentially decayed weights. The solid \textcolor{keywords}{blue} line denotes the training dynamic of JSD between SBT and \textsc{fixed-ripple}, while the dashed \textcolor{strings}{red} line denotes the JSD between the uniform distribution and \textsc{fixed-ripple}.}
\label{fig:jsd}
\vskip -0.15in
\end{figure}

To further investigate how spatial weights generated from our proposed stick-breaking transform (SBT;\S\ref{ssec:spatial_weights}) deviate from \textsc{fixed-ripple}, we plot the training dynamic of the average Jensen-Shannon divergence (JSD) between distributions induced by SBT and \textsc{fixed-ripple}, which is shown in Figure~\ref{fig:jsd}. The JSD scores are averaged over all training samples, for each of which we further average over all attention blocks and heads. Intuitively, a higher JSD value reflects a large discrepancy between their induced distributions. Since the logits in SBT are usually initialized around 0, the spatial weights are close to the exponential weights during the early stage of training; however, as soon as the training starts, the JSD value rises sharply, which is possibly due to balancing between global and local information; after that, the curve decreases slightly, indicating that the mechanism might tend to favor local correlations; finally it plateaus at a high JSD value, which indicates that the induced distribution does not simply degenerate to a fixed distribution nor to a vanilla linearized attention (with uniform weights).

\paragraph{On the Effect of Global and Local Information.}
To demonstrate the relation between global and local information, we design another baseline \textsc{truncated-ripple}, which puts a hard termination of rippling process such that all distant groups beyond $R_{\text{max}}=4$ are discarded (i.e., $\alpha_{r}(i,j) = 0, \text{ if } r \geq 4$) instead of merged. This results in a limited receptive field without global dependency modeling. As shown in Table~\ref{tb:analysis-distance}, the comparison among \textsc{truncated-ripple}, \model and \textsc{deit-la} reveals that both global and local information play an important role in modeling, while the notion of locality is possibly more important than global connectivity in vision tasks, which concurs with previous findings \citep{vit,convit}.

\subsection{Alternative Partitioning schemes for Vicinal Groups}
\label{ssec:log_rippling}

Ripple attention is a flexible framework in balancing between the running time complexity and predictive accuracy. To explore this, we compare the full ripple attention against a rippling process where ripples get exponentially thicker so that the process could reach the image boundary in \emph{logarithmic} time. Formally, recall that vicinal groups with respect to a query token at position $(i,j)$ is defined as  $\{\mathcal{N}_r(i,j)\}_{r=0}^R$, where every token at position $(m,n)$ belongs to $\mathcal{N}_r(i,j)$ if and only if $\max(|m - i|, |n-j|) = r$. We generalize this notion by relaxing the equality condition, that is, $(m,n) \in \mathcal{N}_r(i,j)$ if $2^r \leq \max(|m - i|, |n-j|) < 2^{r+1}$, which allows the size of vicinal groups to be exponentially larger instead of keeping constant. In this way, the number of vicinal groups $R'$ is $\mathcal{O}(\log R)$. This method, which we refer to as \textsc{logarithmic-ripple}, enjoys $\mathcal{O}(HW\log R)$ time complexity and becomes more efficient than base ripple attention. As reported in Table~\ref{tb:analysis-distance}, we see \textsc{ripple-logarithmic} also outperforms \textsc{deit-la} by a large margin, although leading to a clear performance drop compared to the full ripple attention. This may be due to that \textsc{ripple-logarithmic} processes visual tokens at a coarser-grained level. Nevertheless, \textsc{ripple-logarithmic} justifies the flexibility of our framework that one could trade off the task accuracy for higher efficiency and vice versa. More details and run-time comparison can be found in \S\ref{app:ssec:spatial-weight-scheme}.

\subsection{Empirical Running Time and Memory Consumption}
\label{ssec:time}
To verify the advantage of asymptotically faster running complexity in \model, we conduct a simulation experiment on vision transformers to compare the empirical running time and memory consumption of \model against its baselines under different numbers of tokens. The detailed setup can be found in Appendix~\ref{app:time-setup}. Figures~\ref{fig:ripple-speed} demonstrate the comparison results. As mentioned in \S\ref{ssec:dp}, both \textsc{deit} and \textsc{ripple} (Na\"ive) come with quadratic complexity in the number of tokens. We observe that \model with dynamic programming (DP) performs significantly better than \textsc{ripple} (Na\"ive), which demonstrates the effectiveness of our dynamic programming algorithm. Furthermore, \model behaves similarly to \textsc{deit-la} as the number of tokens increases, verifying that it could be executed in linear observed time. When processing a large number of tokens, \model often achieves a 5$\times$ or even 10$\times$ reduction in running time and memory compared to its quadratic counterparts. 

\section{Related Work}
\label{sec:related-work}

Transformer architectures \citep{vaswani2017attention} are first introduced for neural machine translation. Recently, researchers begin to apply the transformer model in the computer vision domain, showing promising results in various tasks, such as image generation \citep{igpt}, video action recognition \citep{timesformer, liu2021videoswin}, segmentation \citep{instance-segmentation, segmenter}, object detection \citep{detr, smca}, low-level image processing \citep{IPT} and image classification \citep{vit,deit,swin-transformer}. A large body of research has been devoted into improving efficiency and effectiveness of vision transformers \citep{vit}. Recent advances improve original vision transformers from various perspectives, such as data-efficient training \citep{deit}, adopting pyramid architectures \citep{pvt,swin-transformer,pit} and incorporating the notion of locality, which can be done by applying convolutional modules into the architecture \citep{localvit,cvt,contnet, vitae, ceit, convit}, restricting the scope of the self-attention \citep{swin-transformer, cswin, chen2021regionvit} or initializing self-attention maps as a convolution kernel \citep{convit}. In contrast to these prior works, we directly model the locality inside the attention mechanism yet permit long-term dependencies, without relying on any convolutional operations or limiting the receptive field; at the same time, ripple attention runs in linear observed time so that the quadratic bottleneck in standard vision transformers can be greatly alleviated. Our work is orthogonal to previous works that modify the transformer architecture and it is worth exploring their combination to improve the overall vision transformer model design.

Our model is built on the linearized attention mechanism, which approximates the softmax kernel with the dot product of feature maps. 
The feature maps can be stochastic, such as in RFA \citep{rfa} and Performer \citep{performer}, or deterministic \citep{katharopoulos2020transformers,schlag2021linear}. Recently, many works are proposed to improve linearized attention by incorporating relative positional encodings \citep{spe, luo2021stable, permuteformer, rotary}. Other efficient attention mechanisms include methods that limit the attention pattern to be sparse \citep{sparse-transformer,axial,reformer} or utilizes a low rank approximation by projecting input sequences to fewer key-value pairs \citep{linformer}. A comprehensive review of recent advances in efficient attention mechanisms can be found in \cite{efficient-attn-1, efficient-attn-2}.

\section{Conclusion}
\label{sec:conclusion}
In this work, we present ripple attention, a novel attention mechanism for visual perception with sub-quadratic complexity. In ripple attention, contributions of different tokens to a query are weighted with respect to their spatial distances in the 2D space.  We design a dynamic programming algorithm that computes weighted contributions for all queries in linear observed time and derive the spatial weights through an adaptive stick-breaking transformation. We conduct extensive experiments and analyses to demonstrate the effectiveness of ripple attention.

\section*{Acknowledgements}
We thank the anonymous reviewers for their valuable suggestions that greatly helped improve this work. This research was supported in part by the joint research scheme of the National Natural Science Foundation of China (NSFC) and the Research Grants Council (RGC) under grant number N\_HKU714/21.

\bibliography{refs}
\bibliographystyle{icml2022}

\newpage
\appendix
\onecolumn
\appendixhead
\appendix
\section{Analysis of Runtime Complexity}
\label{app:sec:complexity}
In this section, we give more details about the runtime complexity (\S\ref{ssec:dp}) for different attention mechanisms in Table~\ref{tb:complexity}. In particular, we focus on variants (1) Ripple-softmax and (2) Ripple (na\"ive), since the rest have been clarified in the main text.

\paragraph{Complexity of Ripple-softmax.}
Ripple-softmax aims to inject radial bias through rippling into the vanilla softmax attention. Here we present two possible ways to achieve this goal, either explicitly or implicitly. Given a query, we could perform vanilla attention over one of its vicinal groups once so that the rippling effect could be explicitly encoded. Since the number of tokens within vicinal group of distance $r$ is directly proportional to $r$ (note that the number of tokens within vicinal group of distance $r$ is simply the difference between $(2r+1)^2 - (2r-1)^2 = 8r$) and there are $R$ groups in total, the overall complexity would be $\mathcal{O}(R^2)$; then considering all queries results in $\mathcal{O}(HWR^2)$ complexity (along with a \textbf{large} constant). On the other hand, we could also add specific spatial weights directly to the attention matrix, implicitly enforcing the radial bias. This is similar to relative positional encodings \citep{shaw2018self}, but in this case the overall complexity is at least $\mathcal{O}(H^2W^2)$ due to the computation of attention matrices, which is as inefficient as vanilla softmax attention in terms of complexity. In this work, we simply refer to the explicit method as Ripple-softmax.

\paragraph{Complexity of Ripple (na\"ive).}
Na\"ively implementing ripple attention comes with $\mathcal{O}(HWR^2)$ complexity, as mentioned in the beginning of $\S~\ref{ssec:dp}$. This is due to the fact that for each query, we have to first sum over all tokens in one vicinal group, which is $\mathcal{O}(r)$ (see the analysis in the paragraph above), and then aggregate over all groups. This results in $\mathcal{O}(R^2)$ complexity, and $\mathcal{O}(HWR^2)$ for all queries (also with a large constant; see the empirical running comparison in \S\ref{ssec:time}).

\section{Efficient Gradient Computation in Ripple Attention}
\label{app:sec:gradient}
To perform gradient back-propagation for ripple attention, a naive implementation would be directly adopting the automatic differentiation, since all operations in our forward pass (Algorithm~\ref{alg:ripple}) are differentiable; however, since many computations in Algorithm~\ref{alg:ripple} overlap with each other, it would lead to substantially repetitive calculations. In addition, we find it even takes $\mathcal{O}(T^2)$ time and memory, which is highly inefficient compared to the forward pass.

In this section, we present an algorithm based on dynamic programming to perform efficient back-propagation for ripple attention, which again comes with sub-quadratic complexity. Recall that given a single query $\vq_{ij}$, all the keys $\mK$ and values $\mV$, ripple attention executes the following computation during the forward pass\footnote{Without loss of generality, we merge the feature map $\phi(\cdot)$ into the vector representation of queries and keys to simplify notations.}:
\begin{align*}
  \operatorname{RippleAttn}\left(\vq_{ij},\mK,\mV\right) \coloneqq 
  \frac
  {\vq_{ij}^\top\sum_{r=0}^R \alpha_{r}(i,j)\sum_{(m,n) \in \mathcal{N}_r(i,j)}\vk_{mn}\vv_{mn}^{\top}}
  {\vq_{ij}^\top\sum_{r=0}^R \alpha_{r}(i,j)\sum_{(m',n') \in \mathcal{N}_r(i,j)} \vk_{m'n'}} \numberthis{}\label{supp-eqn:ripple}.
\end{align*}
The main difference between linearized attention and ripple attention lies in the computation procedure of summation over $\vk\vv^{\top}$ and $\vk$. Therefore, we put main focus on calculating gradients for the following quantity
\begin{equation}
  \vy_{ij} \coloneqq \sum_{r=0}^R \alpha_{r}(i,j)\sum\nolimits_{(m,n) \in \mathcal{N}_r(i,j)}\vx_{mn}, \label{supp-eqn:general}
\end{equation}
where $\vx_{mn} \in \mathbb{R}^D$ denotes a $D$-dimensional vector located at position $(m,n)$, which could be (unrolled) $\vk_{mn}\vv_{mn}^{\top}$ or $\vk_{mn}$.\footnote{We focus on the general form here since the derived algorithm applies to both the nominator and the denominator.}  The remaining computation in ripple attention (e.g., dot product with $\vq_{ij}$) can be easily handled by standard automatic differentiation. We are mainly interested in computing gradients with respect to $\alpha_{r}(i,j)$ and $\vx_{mn}$

In ripple attention, we maintain a summed area table (SAT) to efficiently retrieve the \emph{partial} reduction over vicinal groups (see Algorithm \ref{alg:ripple} for more details).
Although all operations in Algorithm \ref{alg:ripple} is differentiable and thus admits the use of automatic differentiation to calculate gradients, it is very inefficient since computations of most intermediate nodes in the computation graph (for example, $\mS_1$, $\mS_2$, $\mS_3$ and $\mS_4$ in Algorithm \ref{alg:ripple}) overlap with each other, resulting in a large amount of repeated computation.

Here we inspect the form (\eqref{supp-eqn:general}) and show that the properties of vicinal groups could be made use of to derive efficient gradient computation.
\paragraph{Gradients with respect to spatial weights.}
We assume gradients $\nabla_{\vy_{ij}} \mathcal{L}$, that is, the gradient of our loss objective w.r.t. output at all positions $(i,j)$ are available during back-propagation. According to the chain rule, the partial derivative of the objective $\mathcal{L}$ w.r.t. $\alpha_{r}(i,j)$ has the following form:
\begin{align*}
    \frac{\partial \mathcal{L}}{\partial \alpha_{r}(i,j)} = \sum_{d=1}^D \frac{\partial \mathcal{L}}{\partial \evy_{ijd}} \frac{\partial \evy_{ijd}}{\partial \alpha_{r}(i,j)} = \sum_{d=1}^D\frac{\partial \mathcal{L}}{\partial \evy_{ijd}}\sum\nolimits_{(m,n) \in \mathcal{N}_r(i,j)}\evx_{mnd},
\end{align*}
where $\evx_{mnd}$ and $\evy_{mnd}$ denote the $d$-th dimension of $\vx_{mn}$ and the output $\vy_{mn}$ respectively. The first quality holds since spatial weights at every position only depends on the output at that position; but since the same spatial weight applies to all dimensions of $\vy_{ij}$, we have to reduce over the embedding dimension to compute the partial derivative. Similar to the forward pass computation (\eqref{eqn:ripple_sat}), we recognize that the inner summation over the vicinal group $\mathcal{N}_r(i,j)$ can be again computed efficiently by utilizing SATs with Algorithm~\ref{alg:ripple}.

\paragraph{Gradients with respect to $\vx_{mn}$.}
The partial derivative w.r.t. element $\evx_{mnd}$ can be written as
\begin{align*}
    \frac{\partial \mathcal{L}}{\partial \evx_{mnd}} &= \sum_{i=1}^H \sum_{j=1}^W \frac{\partial \mathcal{L}}{\partial \evy_{ijd}} \frac{\partial \evy_{ijd}}{\partial \evx_{mnd}} \\
    &= \sum_{i=1}^H \sum_{j=1}^W \frac{\partial \mathcal{L}}{\partial \evy_{ijd}}\sum_{r=0}^R \alpha_{r}(i,j)\mathbb{I}\left[(m,n) \in \mathcal{N}_r(i,j)\right]. \numberthis{} \label{supp-eqn:grad-x}
\end{align*}
where we define $\mathbb{I}\left[(m,n) \in \mathcal{N}_r(i,j)\right]$ as the indicator function such that it is set to 1 if $(m,n) \in \mathcal{N}_r(i,j)$ and 0 otherwise. A naive way to compute the partial derivatives above has $\mathcal{O}(H^2W^2R)$ complexity, since for every key vector at position $(m,n)$ we need to sum its influences over all positions. However, we show that we could again solve them via dynamic programming.

Our key observation is that the vicinal group is symmetrical w.r.t. its arguments, that is, $(m,n) \in \mathcal{N}_r(i,j)$ if and only if $(i,j) \in \mathcal{N}_r(m,n)$. Then the partial derivative (\eqref{supp-eqn:grad-x}) is equivalent to
\begin{align*}
    \frac{\partial \mathcal{L}}{\partial \evx_{mnd}}
    &= \sum_{i=1}^H \sum_{j=1}^W \frac{\partial \mathcal{L}}{\partial \evy_{ijd}}\sum_{r=0}^R \alpha_{r}(i,j)\mathbb{I}\left[(m,n) \in \mathcal{N}_r(i,j)\right] \\
    &= \sum_{i=1}^H \sum_{j=1}^W \frac{\partial \mathcal{L}}{\partial \evy_{ijd}}\sum_{r=0}^R \alpha_{r}(i,j)\mathbb{I}\left[(i,j) \in \mathcal{N}_r(m,n)\right]\\
    &= \sum_{r=0}^R \sum_{i=1}^H \sum_{j=1}^W \mathbb{I}\left[(i,j) \in \mathcal{N}_r(m,n)\right]\frac{\partial \mathcal{L}}{\partial \evy_{ijd}} \alpha_{r}(i,j)\\
    &= \sum_{r=0}^R \sum\nolimits_{(i,j) \in \mathcal{N}_r(m,n)} \frac{\partial \mathcal{L}}{\partial \evy_{ijd}} \alpha_{r}(i,j).
\end{align*}
Thanks to the symmetry, the computation of partial derivatives is converted into reduction over vicinal groups, it can be effectively solved by dynamic programming (\S\ref{ssec:dp}) in again $\mathcal{O}(HWR)$ time, which involves instantiating an SAT for the quantity $\frac{\partial \mathcal{L}}{\partial \widetilde{\evy}_{ijcd}} \alpha_{r}(i,j)$ over all positions $(i,j)$. Equipped with this result, substituting  $\vk_{mn}\vv_{mn}^{\top}$ or $\vk_{mn}$ into $\vx_{mn}$ yields the term in the nominator and denominator, respectively.

\begin{table}[tb]
	\caption{Runtime complexity comparisons between different attention variants, with respect to an image with $H \times W$ patches (in the first row) and with respect to the number of patch tokens $T \coloneqq H \times W$ (in the second row).
	\textsuperscript{\textdagger}Ripple-softmax indicates the complexity if we would like to implement ripple-like mechanisms in vanilla softmax attention.}
	\label{tb:complexity}
	\vskip 0.1in
	\centering
		\begin{tabular}{c|c|c|c|c}
			\toprule[.1em]
	    {Softmax attention} & {Ripple-softmax\textsuperscript{\textdagger}} & {Linearized attention} & {Ripple (na\"ive)} & \textbf{Ripple (DP)} \\
		\midrule
        $\mathcal{O}\left(H^2W^2\right) $&  $\mathcal{O}\left(HWR^2\right)$ & $\mathcal{O}\left(HW\right)$& $\mathcal{O}\left(HWR^2\right)$ & $\mathcal{O}\left(HWR\right)$  \\
        $\mathcal{O}\left(T^2\right) $&  $\mathcal{O}\left(T^2\right)$ & $\mathcal{O}\left(T\right)$& $\mathcal{O}\left(T^2\right)$ & $\mathcal{O}\left(T^{3/2}\right)$  \\
			\bottomrule[.1em]
	\end{tabular}
	\vskip -0.1in
\end{table}

\section{Additional Implementation Details}
\label{app:sec:implementation}
We implement our model using PyTorch \citep{pytorch} and PyTorch image models (\texttt{timm}) toolkit \citep{timm}. We also implement a CUDA kernel for the ripple attention mechanism. 

\subsection{Deterministic Adaptive Feature Maps for Linearized Attention}
\label{app:ssec:rfa}

\paragraph{Background.}
Generally, a random feature map $\phi_{\mathbf{\omega}}(\cdot)$ is defined by a function $h(\cdot) : \R^D \rightarrow \R$, $m$ uni-variate functions $f_1, f_2, \dots, f_m : \R \rightarrow \R$ as well as $d$ identically distributed random vectors $\omega_1, \omega_2, \dots, \omega_d$ following some distribution \citep{performer}:
\begin{equation*}
    \phi_{\mathbf{\omega}}(\vx) \coloneqq \frac{h(x)}{\sqrt{d}} \left[f_1(\omega_1^\top \vx), \dots, f_1(\omega_d^\top \vx), \dots, f_m(\omega_1^\top \vx), \dots, f_m(\omega_d^\top \vx)\right]
\end{equation*}
yielding a map from $\R^D$ to $\R^{D'}$, where $D' = md$. Then by setting different configurations of $f$'s, $\omega$'s and $h$, we could construct various unbiased estimators for the quantity $\exp(\vx^\top \vy)$, that is, 
\begin{align*}
  \exp(\vx^\top \vy)=\E_{\omega_{1},\dots,\omega_{d}}\left[\phi_{\mathbf{\omega}}(\vx)^\top\phi_{\mathbf{\omega}}(\vy)\right]
\end{align*}
For instance, we could let $m = 2$, where $f_1 = \sin$, $f_2 = \cos$ are trigonometric functions and $h(\vx) = \exp\left(||x||^2 / 2\right)$ \citep{rfa,performer}. Although unbiased, researchers note that the use of trigonometric functions does not ensure non-negative scores, which may lead to large estimate variance and unstable training \citep{performer}. Alternatively, we could construct an estimator by setting $m = 1$ with $f_1 = \exp$ and $h(\vx) = \exp\left(-||x||^2 / 2\right)$, which is again unbiased but enjoys positiveness \citep[FAVOR+,][]{performer}.

\paragraph{Our proposed deterministic adaptive feature map.}
Recently, researchers also proposed various heuristic designs of feature maps \citep{performer, schlag2021linear, kasai2021finetuning} that do not guarantee unbiasedness but might either exhibit lower variance, simplify computation or bring other useful benefits. Unfortunately, through extensive preliminary experiments we found most of these linearized attention variants (either random or deterministic) did not work well in the setting of vision transformers. We hypothesize there are two reasons for the performance drop: the first one is the usage of random samples, which suffers from the slow Monte Carlo convergence rate and instability during training; the second one is due to fixed weights, preventing the map from being adaptive and learning useful patterns. To this end, we propose the following deterministic feature map:
\begin{equation}
    \phi(\vx) = \operatorname{ReLU}(\mW_2 [\sin(\mW_1 \vx); \cos(\mW_1 \vx)] + \vb_2).
\end{equation}
Intuitively, we still follow the trigonometric feature map, except that we set $\mW_1$ to be initialized as independent standard Gaussian samples but then learnable during training; the generated feature is then passed through a fully connected layer followed by a ReLU activation. It is deterministic and involves learnable parameters, which we found greatly improves performance.

\paragraph{Comparison with other feature maps and ablation study.}
We conduct a simple ablation study to demonstrate the effectiveness of our proposed feature map and report comparisons with other feature maps\footnote{
For methods that adopt random features, we sample a set of random weights at every training step and use the same set of weights during evaluation. We also attempted various ways to schedule the redrawing random weights during training, but did not observe any performance gain.}, as shown in Table~\ref{tb:analysis-different-map}. In general, we find it works pretty well in practice and outperforms other feature maps that are either deterministic or random. For our ablation study, we consider two variants of our proposed approach: (1) the method that recovers the original random trigonometric feature map, that is, recasting $\mW_1$ as random samples and re-drawing it at every iteration; (2) the method that removes the fully connected layer (characterized by parameters $\mW_2$ and $\vb_2$). From Table~\ref{tb:analysis-different-map}, we see a great performance drop if we use random weights, which indicates that random feature maps lead to more difficult training in vision transformers. In addition, a feed-forward layer will give a further performance boost due to the increased flexibility. Therefore, we adopt our proposed deterministic feature map throughout our work.

\begin{table}
	\caption{Classification results on \imagenet dataset under different choices of feature maps. \textsuperscript{\textdagger} indicates that our feature map design without fully connected network is identical to T2R \citep{kasai2021finetuning}. \textsuperscript{*} denotes the model does not fully converge.}
	\label{tb:analysis-different-map}
	\vskip 0.1in
	\centering
		\begin{tabular}{c | c | c}
			\toprule[.1em]
			Feature map & Deterministic & Top-1 Acc.\\
			\midrule
    		RFA \citep{rfa} & \xmark & 67.10\textsuperscript{\:\:}       \\
    		Performer \citep{performer}  & \xmark & 65.92\textsuperscript{\:\:}  \\
    		DPFP \citep{schlag2021linear}   & \cmark  & 63.95\textsuperscript{*}  \\
    		T2R \citep{kasai2021finetuning} & \cmark & 70.02\textsuperscript{\:\:} \\
    		\midrule
    		Ours   & \cmark & \textbf{70.67}\textsuperscript{\:\:}  \\
    		\midrule
    		Ours w/ randomly sampled $\mW_1$  & \xmark & 66.82\textsuperscript{\:\:}  \\
    		Ours w/o fully connected network   &  \cmark & 70.02\textsuperscript{\textdagger}  \\
			\bottomrule[.1em]
	\end{tabular}
	\vskip -0.1in
\end{table}

\subsection{Explicitly Controlling the Maximum Rippling Distance}
\label{app:ssec:max_ripple_distance}
In \S\ref{ssec:spatial_weights} we define the threshold $\tau$ to control the termination of rippling process. In practice we find it beneficial to introduce a hard constraint $R_{\text{max}}$ such that the model limits the maximum distance of rippling propagation to $R_{\text{max}}$ and then merges all the remaining groups. In this way, we could not only further reduce the computation overhead, but also encourage the attention mechanism to allocate more weights to distal groups. This can be seen as a stronger version of halting threshold $\tau$, which is easier to tune due to a more intuitive effect on the rippling process. We find an intermediate value gives a reasonable trade-off between local and long-term dependencies. Given $R_{\text{max}}$, our model is robust to the change of $\tau$; therefore, we only set $\tau$ to 0.001 by default and mainly conduct ablation studies on $R_{\text{max}}$.

\subsection{Parameterization of Spatial Weights}
\label{app:ssec:spatial_weights_param}
In terms of parameterizing spatial weights, we allocate an embedding vector for every of $R_{\text{max}}$ stick units, so that they could adapt themselves to learn useful patterns from data. To compute spatial weights, we first linearly project each value vector $\vv_{ij}$ and then perform dot-product with each of $R_{\text{max}}$ stick unit embeddings\footnote{Since stick-breaking transformations ensure the produced weights to be inside a simplex, $R_{\text{max}}$ logits would suffice to produce a sequence of spatial weights with size $R_{\text{max}}+1$.} to produce $R_{\text{max}}$ logits $\{\ro_r(i,j)\}_{r=1}^{R_{\text{max}}}$. Every logit is then passed through a modified sigmoid function to yield the length of each stick unit $\rs_r(i,j) = 1/\left[1 + (R_{\text{max}}-r)\exp{(-\ro_r(i,j))}\right]$. This modification, which is inspired by the default stick-breaking transform implementation in PyTorch distribution package \citep{pytorch}, ensures the model does not put most of mass on the first several sticks. We find this trick slightly improves performance. Consequently, spatial weights $\{\alpha_r(i,j)\}_{r=0}^{R_{\text{max}}}$ are derived by applying stick-breaking transformations to $\rs_r(i,j)$'s according to \eqref{eqn:sbt}.

\subsection{Architecture Details}
\label{app:ssec:arch}
For image classification, all model architectures follow the tiny variant of \textsc{deit} \citep{deit}, consists of 12 transformer layers, with the embedding dimension set to 192, except that we set the number of heads per attention block to 6 for all models. For object detection, our model is based on the architecture of \textsc{smca} with single scale features \citep{smca}, which could facilitate comparisons and demonstrate the effectiveness of ripple attention more clearly. In particular, the number of transformer layers is 6 for both the encoder and decoder, with the number attention heads and the embedding dimension set to 8 and 256, respectively; the backbone is the pre-trained ResNet-50 \citep{resnet} on \imagenet with fixed batch-norm layers. 

\subsection{Specifics for Applying Ripple Attention in Vision Transformers}
\paragraph{Average pooling instead of using class tokens for classification.} Since ripple attention directly operates on 2D images, it is hard to directly employ the widely used class token for classification tasks \citep{vit,deit}.
Instead, we adopt mean average pooling over all tokens instead of class tokens to extract feature vectors that are fed into the classification head.

\paragraph{Multi-head ripple attention.} 
Similar to multi-head attention \citep{vaswani2017attention}, which is used in most Vision transformer architectures, we also adopt a multi-head variant of ripple attention, where different heads maintain different sets of spatial weights. The multi-head ripple attention allows different heads to focus on locality to various degrees, increasing the overall expressiveness.

\paragraph{On the number of ripple layers.} 
A straightforward implementation choice is to replace regular attention at all layers of ViT with ripple attention. However, we find empirically only replacing several initial transformer layers works equally well. Since the input tokens of transformers consist of local patches, promoting local correlations at lower layers and maintaining structural spatial contexts could facilitate information aggregation; but as tokens go higher, every token is contextualized by global information and in this case adding the notion of locality might mislead the modeling. Therefore, we propose to use a hybrid architecture, where the lower layers use ripple attention while upper ones still adopt linear attention mechanisms. This choice is further supported by our ablation study experiments Appendix~\ref{app:ssec:ripple-layers}, where our model achieves the best performance over various settings if only the first 9 transformer layers use ripple attention. Therefore, throughout experiments we use this configuration unless otherwise stated.

\subsection{Training Setup}
\label{app:ssec:training_details}
In this section, we describe our full training setup for both image classification and object detection.

\paragraph{Training details for image classification}
We following the same procedure to train the models as in \textsc{deit} \citep{deit}, including the data-augmentation, the regularization and the hyper-parameter setting for a head-to-head comparison.
We use AdamW optimizer \citep{adamw} to train our model on 8 NVIDIA V100 GPUs for 300 epochs on both \cifar and \imagenet datasets. We adopt commonly used data augmentation methods, including random clipping, cropping, Rand-Augment \citep{random-augment} and random erasing \citep{random-erasing}. However, we remove repeated augmentation \citep{repeat-augment} as we find it slows down convergence for both linearized attention and ripple attention, as also observed in previous studies \citep{berman2019multigrain,xiao2021early}. For regularization, we employ stochastic depth \citep{stochastic-depth}, Mixup \citep{mixup}, Cutmix \citep{cutmix}, all of which are set to default settings in DeiT \citep{deit}. Training protocols that are specific to different datasets are listed as follows:
\begin{itemizesquish}
    \item For \imagenet dataset we set the batch size to 1024 and the learning rate to 0.001 with cosine learning rate decay \citep{cos-lr}. The image size is set to $224 \times 224$ with patch size $16$, resulting in $14 \times 14$ tokens.
    \item for \cifar dataset, the batch size and the learning rate is set to 512 and 0.0005 respectively, with the same cosine learning rate decay. In terms of the image size, we use the original scale $32 \times 32$, where a patch size $2$ is used to produce $16 \times 16$ non-overlapping patches. 
\end{itemizesquish}
During evaluation, we report top-1 and top-5 accuracy on the evaluation set of both \imagenet and \cifar datasets.


\paragraph{Training details for object detection}
We follow the same training protocol as SMCA \citep{smca}. In particular, we initialize the transformer parameters with Xavier initialization \citep{xavier}, and use the pretrained weights on \imagenet for the backbone. We adopt the AdamW optimizer \citep{adamw}, set the weight decay to $10^{-4}$ and the learning rate to $10^{-5}$ and $10^{-4}$ for the backbone and transformer, respectively. We also decrease the learning rate to 1/10 of its original value after 40 epochs for 50 epoch schedule and after 80 epochs for 108 epoch training schedule. The dropout rate is set to 0.1. The data augmentation scheme and the loss objective is also the same as SMCA \citep{smca}. All detection models are trained on 8 NVIDIA V100 GPUs with a total batch size of 16.

\section{Additional Experiment Results}
\label{app:sec:more-experiments}

\subsection{On the Effect of Various Ripple Layers}
\label{app:ssec:ripple-layers} 
As mentioned in \S\ref{ssec:setup}, directly replacing all attention layers in \textsc{deit-la} with \model could be a sub-optimal choice. 
To validate this, we conduct an ablation study on \imagenet dataset to investigate the effect of different numbers of ripple layers, where the first several layers use ripple attention while upper ones still adopt linearized attention mechanism. The results are shown in Table~\ref{app:tb:analysis-layer}. In particular, we find the model performance consistently improves as the number of ripple layers increases, but drops a little when the depth of ripple layers reaches a certain level (e.g., 9). Our observation aligns with our intuition, which suggests using hybrid attention layers could achieve a good trade-off between locality promotion and global dependency modeling. Therefore, \model uses 9 ripple layers by default throughout our experiments unless otherwise stated. 

\begin{table}[t]
	\caption{Classification results on \imagenet dataset under different numbers of rippling layers for \model. The speed is measured by the number of images processed per second with a batch size of 64 on a single NVIDIA V100 GPU machine, averaged by 5 runs.}
	\label{app:tb:analysis-layer}
	\vskip 0.1in
	\centering
	\resizebox{0.4\columnwidth}{!}{    
		\begin{tabular}{c | c | c | c}
			\toprule[.1em]
			\# ripple layers & Speed & Top-1 Acc. & Top-5 Acc. \\
			\midrule
    		0   & 2664 & 70.67  & 90.16  \\
    		3   & 1355 & 71.63  & 90.42  \\
    		6   & 916 & 72.41  & 90.32  \\
    		9   & 792 & \textbf{73.02}  & \textbf{91.56}  \\
    		12  & 563 & 72.69  & 91.30  \\
			\bottomrule[.1em]
	\end{tabular}}  
	\vskip -0.1in
\end{table}

\subsection{On the Effect of Different Parameterization Schemes for Vicinal Groups}
\label{app:ssec:spatial-weight-scheme} 
Ripple attention is a flexible framework in that it allows the trade-off between the running time complexity and task accuracy. To explore this, we compare the full ripple attention against a rippling process where ripples get exponentially thicker so that the process could reach the image boundary in logarithmic time. Formally, recall that vicinal groups with respect to a query token at position $(i,j)$ is defined as  $\{\mathcal{N}_r(i,j)\}_{r=0}^R$, where every token at position $(m,n)$ belongs to $\mathcal{N}_r(i,j)$ if and only if $\max(|m - i|, |n-j|) = r$. In the setting of \textsc{ripple-logarithmic}, we generalize this notion by relaxing the equality condition, that is, now $(m,n) \in \mathcal{N}_r(i,j)$ if and only if $2^r \leq \max(|m - i|, |n-j|) < 2^{r+1}$, which allows the size of vicinal groups to be exponentially larger instead of keeping constant. In this way, the number of all vicinal groups $R'$ is $\mathcal{O}(\log R)$. This model, which we refer to as \textsc{ripple-logarithmic}, enjoys $\mathcal{O}(HW\log R)$ time complexity and is more efficient than base ripple attention. 

\begin{figure*}[t]
\centering
\begin{subfigure}[b]{0.43\textwidth} 
  \centering
  {\includegraphics[width=\textwidth]{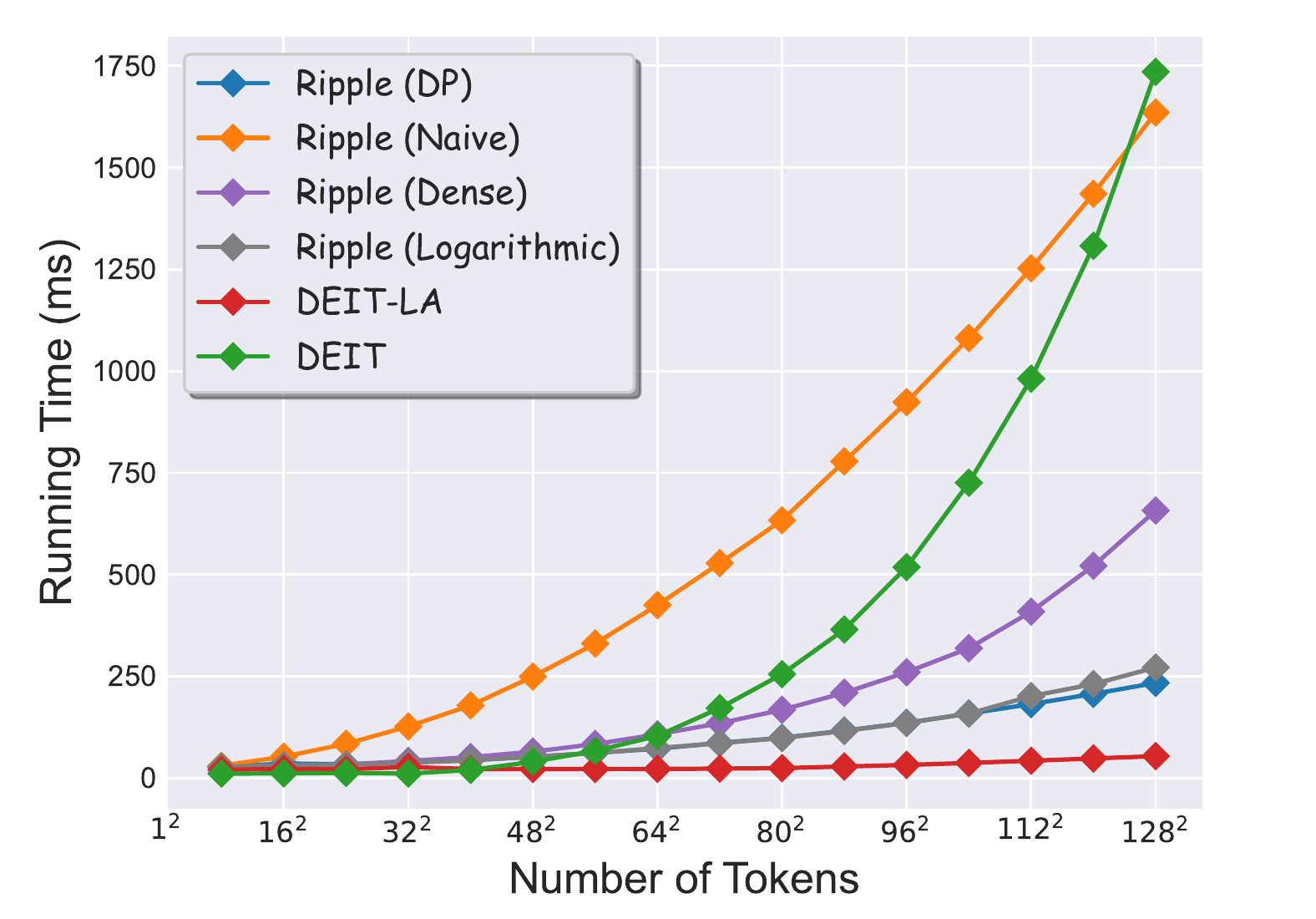}}
\end{subfigure}
\begin{subfigure}[b]{0.43\textwidth} 
  \centering
  {\includegraphics[width=\textwidth]{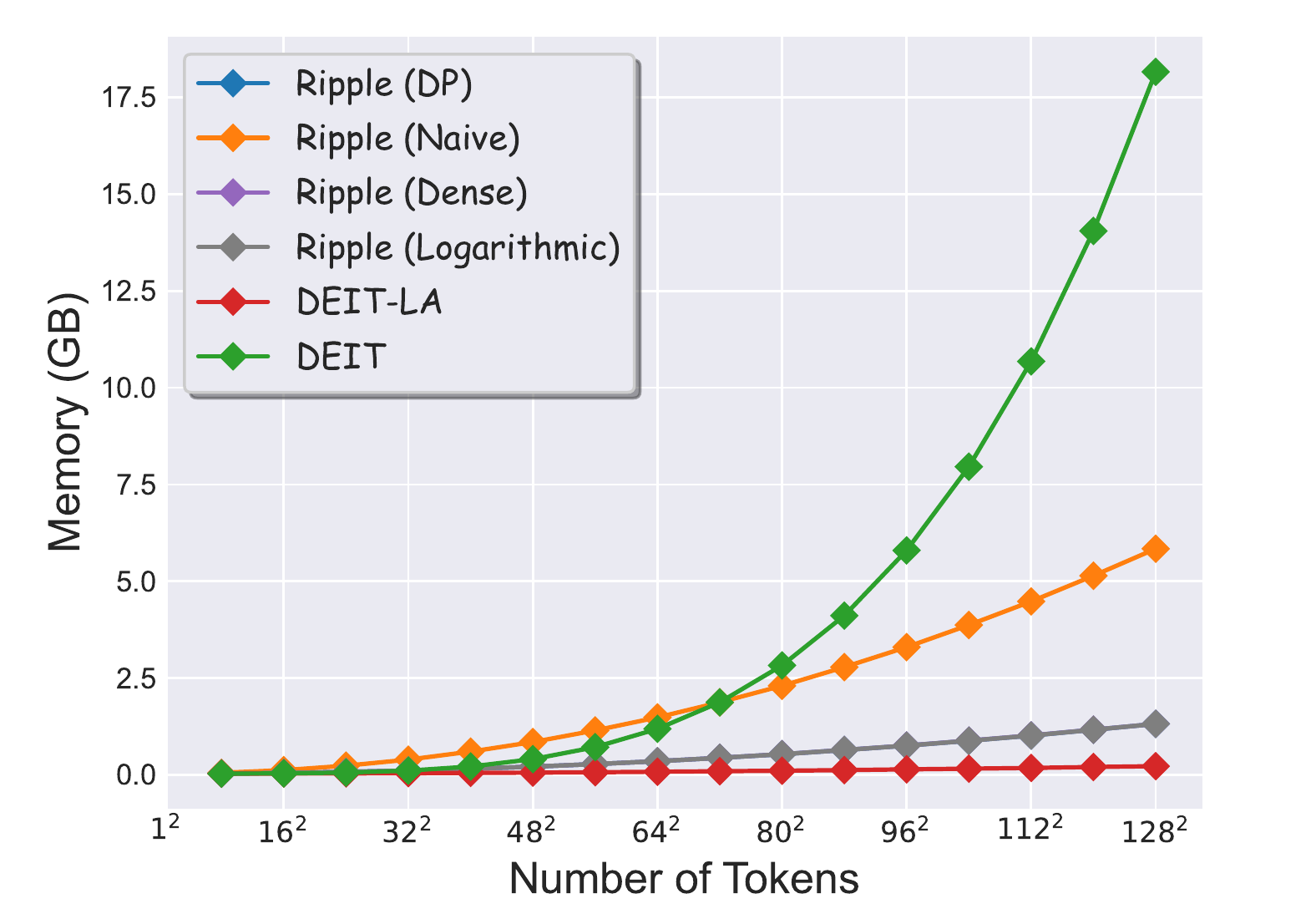}}
\end{subfigure}
\caption{Empirical running time (left) and memory consumption (right) under different numbers of tokens, averaged by 5 runs.}
\label{app:fig:ripple-speed}
\vspace{-5mm}
\end{figure*}

To empirically evaluate the efficiency of this variant, we plot the empirical running statistics of \textsc{ripple-logarithmic} under different numbers of tokens. For completeness, we also include a variant where $R_{\text{max}}$ scales linearly with the image height (or width), denoted by \textsc{ripple-dense}. As shown in Figure~\ref{app:fig:ripple-speed}, we observe \textsc{ripple-logarithmic} runs as fast as base ripple (whose $R_{\text{max}}$ is fixed) and becomes more efficient than the dense version as the number of tokens increases. On the other hand, all of these models run with the same amount of memory consumption, as their space complexity is constant in the rippling distance. In terms of task performance, as reported in Table~\ref{tb:analysis-distance}, we see a clear performance drop if we adopt \textsc{ripple-logarithmic}, which could be due to that \textsc{ripple-logarithmic} processes visual tokens at a coarser-grained level. This again justifies the flexibility of our framework: one could trade off the task accuracy for more efficiency and vice versa. 

\subsection{Performance Comparison under the Same Speed Constraint}
\label{app:ssec:same-speed}
We conduct an ablative experiment to compare the performance of \model against \textsc{deit-la} under the same speed constraint. The hyper-parameter configuration remains the same as the main experiment. As shown in Table~\ref{app:tb:analysis-speed}, we find that a 4-layer transformer with \model outperforms a 12-layer model with \textsc{deit-la} by a reasonable margin while running at a similar speed. This indicates the enlarged modeling capacity of \model compared to traditional linearized attention, demonstrating the effectiveness of our approach.
\begin{table}[t]
\caption{Classification results on \cifar dataset under the same speed constraint.}
\label{app:tb:analysis-speed}
\centering
\resizebox{0.4\columnwidth}{!}{    
\begin{tabular}{l | c | c | c}
\toprule[.1em]
{Models} & {Speed} & {\# Params} & {Top-1 Acc.} \\
\midrule
4-layer \textsc{deit-la} & 6953 & 1.88M & 63.35 \\
\midrule
\textsc{deit-la} & 2686 & 5.50M & 67.10  \\
4-layer \model & 2369 & 1.89M & \textbf{70.03} \\
\midrule
\model  & 893 & 5.52M & \textbf{74.11}\\
\bottomrule[.1em]
\end{tabular}}  
\end{table} 

\section{Setup for Empirical Running Time and Memory Consumption}
\label{app:time-setup}
For the simulation experiment conducted in \S\ref{ssec:time}, we use the same vision transformer architecture for all models, whose hyper-parameter setting is specified in Appendix~\ref{app:ssec:arch}, except that we set embedding dimension to 96 and batch size to 4; otherwise, most configurations tested here will make it infeasible for \textsc{deit} and \textsc{ripple} (Na\"ive) to fit into the 32GB memory of a single NVIDIA V100 GPU machine.

\end{document}